\documentclass[sigconf]{acmart}

\renewcommand\footnotetextcopyrightpermission[1]{} 
\usepackage{graphicx}

\AtBeginDocument{%
  \providecommand\BibTeX{{%
    \normalfont B\kern-0.5em{\scshape i\kern-0.25em b}\kern-0.8em\TeX}}}

\setcopyright{acmcopyright}
\copyrightyear{2018}
\acmYear{2018}
\acmDOI{10.1145/1122445.1122456}

\acmConference[Woodstock '18]{Woodstock '18: ACM Symposium on Neural
  Gaze Detection}{June 03--05, 2018}{Woodstock, NY}
\acmBooktitle{Woodstock '18: ACM Symposium on Neural Gaze Detection,
  June 03--05, 2018, Woodstock, NY}
\acmPrice{15.00}
\acmISBN{978-1-4503-9999-9/18/06}

\begin{document}

\title{A framework for anomaly detection using language modeling, and its applications to finance}

\author{Armineh Nourbakhsh}
\email{armineh.nourbakhsh@spglobal.com}
\affiliation{%
  \institution{S\&P Global Ratings}
  \city{New York}
  \state{NY}
}
\author{Grace Bang}
\email{grace.bang@spglobal.com}
\affiliation{%
  \institution{S\&P Global Ratings}
  \city{New York}
  \state{NY}
}








\renewcommand{\shortauthors}{Nourbakhsh and Bang}

\begin{abstract}
  In the finance sector, studies focused on anomaly detection are often associated with time-series and transactional data analytics. In this paper, we lay out the opportunities for applying anomaly and deviation detection methods to text corpora and challenges associated with them. We argue that language models that use distributional semantics can play a significant role in advancing these studies in novel directions, with new applications in risk identification, predictive modeling, and trend analysis.
\end{abstract}





\maketitle

\section{Introduction}
The detection of anomalous trends in the financial domain has focused largely on fraud detection \cite{fraudfin2016}, risk modeling \cite{risk2018}, and predictive analysis \cite{pred2000}. The data used in the majority of such studies is of time-series, transactional, graph or generally quantitative or structured nature. This belies the critical importance of semi-structured or unstructured text corpora that practitioners in the finance domain derive insights from---corpora such as financial reports, press releases, earnings call transcripts, credit agreements, news articles, customer interaction logs, and social data. 

Previous research in anomaly detection from text has evolved largely independently from financial applications. Unsupervised clustering methods have been applied to documents in order to identify outliers and emerging topics \cite{topic2013}. Deviation analysis has been applied to text in order to identify errors in spelling \cite{ngram2013} and tagging of documents \cite{error2000}. Recent popularity of distributional semantics \cite{distsem2010} has led to further advances in semantic deviation analysis \cite{distsem2011}. However, current research remains largely divorced from specific applications within the domain of finance. 

In the following sections, we enumerate major applications of anomaly detection from text in the financial domain, and contextualize them within current research topics in Natural Language Processing.

\section{Five views on anomaly}
Anomaly detection is a strategy that is often employed in contexts where a deviation from a certain norm is sought to be captured, especially when extreme class imbalance impedes the use of a supervised approach. The implementation of such methods allows for the unveiling of previously hidden or obstructed insights. 

In this section, we lay out five perspectives on how textual anomaly detection can be applied in the context of finance, and how each application opens up opportunities for NLP researchers to apply current research to the financial domain.

\subsection{Anomaly as error}
Previous studies have used anomaly detection to identify and correct errors in text \cite{ngram2013, error2000}. These are often unintentional errors that occur as a result of some form of data transfer, e.g. from audio to text, from image to text, or from one language to another. Such studies have direct applicability to the error-prone process of earnings call or customer call transcription, where audio quality, accents, and domain-specific terms can lead to errors. Consider a scenario where the CEO of a company states in an audio conference, `Now investments will be made in Asia.' However, the system instead transcribes, `No investments will be made in Asia.' There is a meaningful difference in the implication of the two statements that could greatly influence the analysis and future direction of the company. Additionally, with regards to the second scenario, it is highly unlikely that the CEO would make such a strong and negative statement in a public setting thus supporting the use of anomaly detection for error correction.  

Optical-character-recognition from images is another error-prone process with large applicability to finance. Many financial reports and presentations are circulated as image documents that need to undergo OCR in order to be machine-readable. OCR might also be applicable to satellite imagery and other forms of image data that might include important textual content such as a graphical representation of financial data. Errors that result from OCR'd documents can often be fixed using systems that have a robust semantic representation of the target domain. For instance, a model that is trained on financial reports might have encoded awareness that emojis are unlikely to appear in them or that it is unusual for the numeric value of profit to be higher than that of revenue.

\subsection{Anomaly as irregularity}
Anomaly in the semantic space might reflect irregularities that are intentional or emergent, signaling risky behavior or phenomena. A sudden change in the tone and vocabulary of a company's leadership in their earnings calls or financial reports can signal risk. News stories that have abnormal language, or irregular origination or propagation patterns might be unreliable or untrustworthy.

\cite{stability2018} showed that when trained on similar domains or contexts, distributed representations of words are likely to be stable, where stability is measured as the similarity of their nearest neighbors in the distributed space. Such insight can be used to assess anomalies in this sense. As an example, \cite{echochamber2017} identified cliques of users on Twitter who consistently shared news from similar domains. Characterizing these networks as ``echo-chambers,'' they then represented the content shared by these echo-chambers as distributed representations. When certain topics from one echo-chamber began to deviate from similar topics in other echo-chambers, the content was tagged as unreliable. \cite{echochamber2017} showed that this method can be used to improve the performance of standard methods for fake-news detection.

In another study \cite{transparency2017}, the researchers hypothesized that transparent language in earnings calls indicates high expectations for performance in the upcoming quarters, whereas semantic ambiguity can signal a lack of confidence and expected poor performance. By quantifying transparency as the frequent use of numbers, shorter words, and unsophisticated vocabulary, they showed that a change in transparency is associated with a change in future performance.

\subsection{Anomaly as novelty}
Anomaly can indicate a novel event or phenomenon that may or may not be risky. Breaking news stories often emerge as anomalous trends on social media. \cite{novel2017} experimented with this in their effort to detect novel events from Twitter conversations. By representing each event as a real-time cluster of tweets (where each tweet was encoded as a vector), they managed to assess the novelty of the event by comparing its centroid to the centroids of older events. 

Novelty detection can also be used to detect emerging trends on social media, e.g. controversies that engulf various brands often start as small local events that are shared on social media and attract attention over a short period of time. How people respond to these events in early stages of development can be a measure of their veracity or controversiality \cite{rumorbehavior2015, rumorverification2015}.

An anomaly in an industry grouping of companies can also be indicative of a company that is disrupting the norm for that industry and the emergence of a new sector or sub-sector. Often known as trail-blazers, these companies innovate faster than their competitors to meet market demands sometimes even before the consumer is aware of their need. As these companies continually evolve their business lines, their core operations are novel outliers from others in the same industry classification that can serve as meaningful signals of transforming industry demands.

\subsection{Anomaly as semantic richness}
A large portion of text documents that analysts and researchers in the financial sectors consume have a regulatory nature. Annual financial reports, credit agreements, and filings with the U.S. Securities and Exchange Commission (SEC) are some of these types of documents. These documents can be tens or hundreds of pages long, and often include boilerplate language that the readers might need to skip or ignore in order to get to the ``meat'' of the content. Often, the abnormal clauses found in these documents are buried in standard text so as not to attract attention to the unique phrases.

\cite{magnet2017} used smoothed representations of n-grams in SEC filings in order to identify boilerplate and abnormal language. They did so by comparing the probability of each n-gram against the company's previous filings, against other filings in the same sector, and against other filings from companies with similar market cap. The aim was to assist accounting analysts in skipping boilerplate language and focusing their attention on important snippets in these documents.

Similar methods can be applied to credit agreements where covenants and clauses that are too common are often ignored by risk analysts and special attention is paid to clauses that ``stand out'' from similar agreements.

\subsection{Anomaly as contextual relevance}
Certain types of documents include universal as well as context-specific signals. As an example, consider a given company's financial reports. The reports may include standard financial metrics such as total revenue, net sales, net income, etc. In addition to these universal metrics, businesses often report their performance in terms of the performance of their operating segments. These segments can be business divisions, products, services, or regional operations. The segments are often specific to the company or its peers. For example, Apple Inc.'s segments might include ``iPhone,'' ``iMac,'' ``iPad,'' and ``services.'' The same segments will not appear in reports by other businesses. 

For many analysts and researchers, operating segments are a crucial part of exploratory or predictive analysis. They use performance metrics associated with these segments to compare the business to its competitors, to estimate its market share, and to project the overall performance of the business in upcoming quarters. Automating the identification and normalization of these metrics can facilitate more insightful analytical research. Since these segments are often specific to each business, supervised models that are trained on a diverse set of companies cannot capture them without overfitting to certain companies. Instead, these segments can be treated as company-specific anomalies. 

\section{Anomaly detection via language modeling}
Unlike numeric data, text data is not directly machine-readable, and requires some form of transformation as a pre-processing step. In ``bag-of-words'' methods, this transformation can take place by assigning an index number to each word, and representing any block of text as an unordered set of these words. A slightly more sophisticated approach might chain words into continuous ``n-grams'' and represent a block of text as an ordered series of ``n-grams'' that have been extracted on a sliding window of size \textit{n}.  These approaches are conventionally known as ``language modeling.''

\begin{figure}[h]
\centering
\includegraphics[width=8cm]{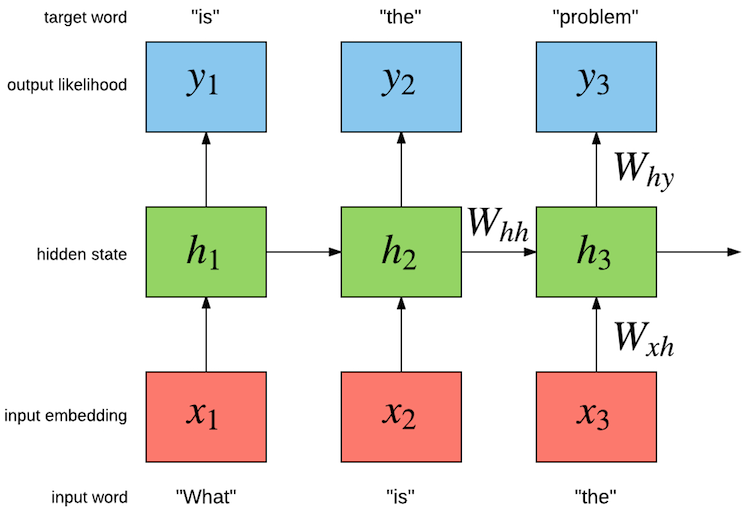}
\caption{Illustration of a recurrent step in a language model. Excerpted from \cite{torch2016}.}
\label{fig:lm}
\end{figure}

Since the advent of high-powered processors enabled the widespread use of distributed representations, language modeling has rapidly evolved and adapted to these new capabilities. Recurrent neural networks can capture an arbitrarily long sequence of text and perform various tasks such as classification or text generation \cite{gen2015}. In this new context, language modeling often refers to training a recurrent network that predicts a word in a given sequence of text \cite{ulmfit2018}. Language models are easy to train because even though they follow a predictive mechanism, they do not need any labeled data, and are thus unsupervised. 

Figure \ref{fig:lm} is a simple illustration of how a neural network that is composed of recurrent units such as Long-Short Term Memory (LSTM) \cite{lstm1997} can perform language modeling. The are four main components to the network:
\begin{itemize}
    \item The input vectors ($x_i$), which represent units  (i.e. characters, words, phrases, sentences, paragraphs, etc.) in the input text. Occasionally, these are represented by one-hot vectors that assign a unique index to each particular input. More commonly, these vectors are adapted from a pre-trained corpus, where distributed representations have been inferred either by a simpler auto-encoding process \cite{word2vec2013} or by applying the same recurrent model to a baseline corpus such as Wikipedia \cite{ulmfit2018}.
    \item The output vectors ($y_i$), which represent the model's prediction of the next word in the sequence. Naturally, they are represented in the same dimensionality as $x_i$s. 
    \item The hidden vectors ($h_i$), which are often randomly initialized and learned through backpropagation. Often trained as dense representations, these vectors tend to display characteristics that indicate semantic richness \cite{elmo2018} and compositionality \cite{word2vec2013}. While the language model can be used as a text-generation mechanism, the hidden vectors are a strong side product that are sometimes extracted and reused as augmented features in other machine learning systems \cite{bert2018}. 
    \item The weights of the network ($W_{ij}$) (or other parameters in the network), which are tuned through backpropagation. These often indicate how each vectors in the input or hidden sequence is utilized to generate the output. These parameters play a big role in the way the output of neural networks are reverse-engineered or explained to the end user \footnote{As an example see \url{https://tinyurl.com/y56drbnk}}.
\end{itemize}

The distributions of any of the above-mentioned components can be studied to mine signals for anomalous behavior in the context of irregularity, error, novelty, semantic richness, or contextual relevance.

\subsection{Anomaly in input vectors}
As previously mentioned, the input vectors to a text-based neural network are often adapted from publicly-available word vector corpora. In simpler architectures, the network is allowed to back-propagate its errors all the way to the input layer, which might cause the input vectors to be modified. This can serve as a signal for anomaly in the semantic distributions between the original vectors and the modified vectors.

Analyzing the stability of word vectors when trained on different iterations can also signal anomalous trends \cite{stability2018}.

\subsection{Anomaly in output vectors}
As previously mentioned, language models generate a probability distribution over a word (or character) in a sequence. These probabilities can be used to detect transcription or character-recognition errors in a domain-friendly manner. When the language model is trained on financial data, domain-specific trends (such as the use of commas and parentheses in financial metrics) can be captured and accounted for by the network, minimizing the rate of false positives.

\subsection{Anomaly in hidden vectors}
\begin{figure*}[h]
\centering
\includegraphics[width=\textwidth]{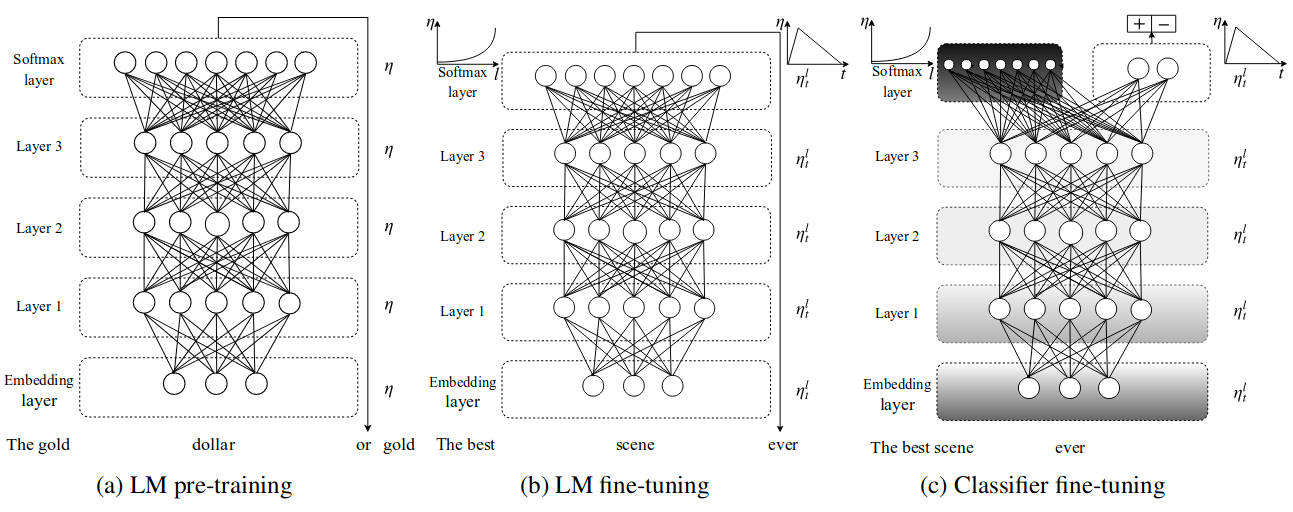}
\caption{A pre-trained model can be fine-tuned on a new domain, and applied to a classification or prediction task. Excerpted from \cite{ulmfit2018}.}
\label{fig:ft}
\end{figure*}

A recent advancement in text processing is the introduction of fine-tuning methods to neural networks trained on text \cite{ulmfit2018}. Fine-tuning is an approach that facilitates the transfer of semantic knowledge from one domain (source) to another domain (target). The source domain is often large and generic, such as web data or the Wikipedia corpus, while the target domain is often specific (e.g. SEC filings). A network is pre-trained on the source corpus such that its hidden representations are enriched. Next, the pre-trained networks is re-trained on the target domain, but this time only the final (or top few) layers are tuned and the parameters in the remaining layers remain ``frozen.'' The top-most layer of the network can be modified to perform a classification, prediction, or generation task in the target domain (see Figure \ref{fig:ft}).

Fine-tuning aims to change the distribution of hidden representations in such a way that important information about the source domain is preserved, while idiosyncrasies of the target domain are captured in an effective manner \cite{transfer2017}. A similar process can be used to determine anomalies in documents. As an example, consider a model that is pre-trained on historical documents from a given sector. If fine-tuning the model on recent documents from the same sector dramatically shifts the representations for certain vectors, this can signal an evolving trend. 

\subsection{Anomaly in weight tensors and other parameters}
Models that have interpretable parameters can be used to identify areas of deviation or anomalous content. Attention mechanisms \cite{attention2017} allow the network to account for certain input signals more than others. The learned attention mechanism can provide insight into potential anomalies in the input. Consider a language model that predicts the social media engagement for a given tweet. Such a model can be used to distinguish between engaging and information-rich content versus clickbait, bot-generated, propagandistic, or promotional content by exposing how, for these categories, engagement is associated with attention to certain distributions of ``trigger words.'' 

\begin{table}[]
\resizebox{\columnwidth}{!}{%
\begin{tabular}{llll}
\textbf{\begin{tabular}[c]{@{}l@{}}Possible \\ source of\\ anomaly\end{tabular}} & \textbf{\begin{tabular}[c]{@{}l@{}}Possible \\ type of \\ anomaly\end{tabular}} & \textbf{\begin{tabular}[c]{@{}l@{}}Example \\ Application\end{tabular}}                                            & \textbf{\begin{tabular}[c]{@{}l@{}}Example\\ Analysis\end{tabular}}                                                 \\ \hline
Input                                                                            & Novelty                                                                         & \begin{tabular}[c]{@{}l@{}}Identifying how \\ perspectives on ESG\\ factors are changing \\ over time in\\financial reports\end{tabular} & \begin{tabular}[c]{@{}l@{}}Retrain the network\\ on y-o-y data and \\ observe unstable \\ word vectors\end{tabular} \\ \hline
Output                                                                           & Error                                                                           & \begin{tabular}[c]{@{}l@{}}Identifying errors\\ in earnings call\\ transcripts\end{tabular}                        & \begin{tabular}[c]{@{}l@{}}Analyze the \\ emission probability \\ of observed\\ words\end{tabular}                  \\ \hline
Hidden                                                                           & \begin{tabular}[c]{@{}l@{}}Semantic\\ richness\end{tabular}                     & \begin{tabular}[c]{@{}l@{}}Identifying\\ non-boilerplate\\ language\end{tabular}                                   & \begin{tabular}[c]{@{}l@{}}Determine which\\ hidden vectors \\ diverge from others\end{tabular}                     \\ \hline
\begin{tabular}[c]{@{}l@{}}Weights \&\\Params\end{tabular}                                                                          & Irregularity                                                                    & \begin{tabular}[c]{@{}l@{}}Identifying\\clickbait\\content\end{tabular}                     & \begin{tabular}[c]{@{}l@{}}Observe how a \\ response-generation\\model attends to\\words in the input \end{tabular}           
\end{tabular}
}
\caption{Four scenarios for anomaly detection on text data using signals from various layers and parameters in a language model.}
\label{tab:apps}
\end{table}

Table \ref{tab:apps} lists four scenarios for using the various layers and parameters of a language model in order to perform anomaly detection from text.


\section{Challenges and Future Research}
Like many other domains, in the financial domain, the application of language models as a measurement for semantic regularity of text bears the challenge of dealing with unseen input. Unseen input can be mistaken for anomaly, especially in systems that are designed for error detection. As an example, a system that is trained to correct errors in an earnings call transcript might treat named entities such as the names of a company's executives, or a recent acquisition, as anomalies. This problem is particularly prominent in fine-tuned language models, which are pre-trained on generic corpora that might not include domain-specific terms. 

When anomalies are of a malicious nature, such as in the case where abnormal clauses are included in credit agreements, the implementation of the anomalous content is adapted to appear normal. Thereby, the task of detecting normal language becomes more difficult. 

Alternatively, in the case of language used by executives in company presentations such as earnings calls, there may be a lot of noise in the data due to the large degree of variability in the personalities and linguistic patterns of various leaders. The noise variability present in this content could be similar to actual anomalies, hence making it difficult to identify true anomalies. 
 
Factors related to market interactions and competitive behavior can also impact the effectiveness of anomaly-detection models. In detecting the emergence of a new industry sector, it may be challenging for a system to detect novelty when a collection of companies, rather than a single company, behave in an anomalous way. The former may be the more common real-world scenario as companies closely monitor and mimic the innovations of their competitors. The exact notion of anomaly can also vary based on the sector and point in time. For example, in the technology sector, the norm in today's world is one of continuous innovation and technological advancements. 
 
Additionally, certain types of anomaly can interact and make it difficult for systems to distinguish between them. As an example, a system that is trained to identify the operating segments of a company tends to distinguish between information that is specific to the company, and information that is common across different companies. As a result, it might identify the names of the company's board of directors or its office locations as its operating segments. 

Traditional machine learning models have previously tackled the above challenges, and solutions are likely to emerge in the neural paradigms as well. Any future research in these directions will have to account for the impact of such solutions on the reliability and explainability of the resulting models and their robustness against adversarial data.

\section{Conclusion}
Anomaly detection from text can have numerous applications in finance, including risk detection, predictive analysis, error correction, and peer detection. We have outlined various perspectives on how anomaly can be interpreted in the context of finance, and corresponding views on how language modeling can be used to detect such aspects of anomalous content. We hope that this paper lays the groundwork for establishing a framework for understanding the opportunities and risks associated with these methods when applied in the financial domain.

\bibliographystyle{ACM-Reference-Format}
\bibliography{references}

\end{document}